\PassOptionsToPackage{table}{xcolor}

\documentclass[sigconf]{acmart}
\usepackage{subcaption}

\usepackage{algorithm}
\usepackage{algpseudocode}
\AtBeginDocument{%
  }

\setcopyright{acmlicensed}
\copyrightyear{2025}
\acmYear{2025}
\acmDOI{XXXXXXX.XXXXXXX}
\acmConference[Conference acronym 'XX]{Make sure to enter the correct
  conference title from your rights confirmation email}{June 03--05,
  2025}{Woodstock, NY}
\acmISBN{978-1-4503-XXXX-X/2025/04}

\acmSubmissionID{5041}

\begin{document}

\title{AeroLite: Tag-Guided Lightweight Generation of Aerial Image Captions}

%
\author{Xing Zi}
\affiliation{%
  \institution{University of Technology Sydney}
  \city{Sydney}
  \country{Australia}
}
\email{xing.zi-1@uts.edu.au}

\author{Tengjun Ni}
\affiliation{%
  \institution{University of New South Wales}
  \city{Sydney}
  \country{Australia}
}
\email{nitengjun2002@gmail.com}

\author{Xianjing Fan}
\affiliation{%
  \institution{University of Technology Sydney}
  \city{Sydney}
  \country{Australia}
}
\email{Xianjing.Fan@student.uts.edu.au}

\author{Xian Tao}
\affiliation{%
  \institution{Institute of Automation, Chinese Academy of Sciences}
  \city{Beijing}
  \country{China}
}
\email{taoxian2013@ia.ac.cn}

\author{Jun Li}
\affiliation{%
  \institution{University of Technology Sydney}
  \city{Sydney}
  \country{Australia}
}
\email{jun.li@uts.edu.au}

\author{Ali Braytee}
\affiliation{%
  \institution{University of Technology Sydney}
  \city{Sydney}
  \country{Australia}
}
\email{Ali.Braytee@uts.edu.au}

\author{Mukesh Prasad}
\affiliation{%
  \institution{University of Technology Sydney}
  \city{Sydney}
  \country{Australia}
}
\email{mukesh.prasad@uts.edu.au}

\renewcommand{\shortauthors}{Zi et al.}

%

\begin{abstract}
Accurate and automated captioning of aerial imagery is crucial for applications like environmental monitoring, urban planning, and disaster management. However, this task remains challenging due to complex spatial semantics and domain variability. To address these issues, we introduce \textbf{AeroLite}, a lightweight, tag-guided captioning framework designed to equip small-scale language models (1--3B parameters) with robust and interpretable captioning capabilities specifically for remote sensing images. \textbf{AeroLite} leverages GPT-4o to generate a large-scale, semantically rich pseudo-caption dataset by integrating multiple remote sensing benchmarks, including DLRSD, iSAID, LoveDA, WHU, and RSSCN7. To explicitly capture key semantic elements such as orientation and land-use types, AeroLite employs natural language processing techniques to extract relevant semantic tags. These tags are then learned by a dedicated multi-label CLIP encoder, ensuring precise semantic predictions. To effectively fuse visual and semantic information, we propose a novel bridging multilayer perceptron (MLP) architecture, aligning semantic tags with visual embeddings while maintaining minimal computational overhead. AeroLite's flexible design also enables seamless integration with various pretrained large language models. We adopt a two-stage LoRA-based training approach: the initial stage leverages our pseudo-caption dataset to capture broad remote sensing semantics, followed by fine-tuning on smaller, curated datasets like UCM and Sydney Captions to refine domain-specific alignment. Experimental evaluations demonstrate that AeroLite surpasses significantly larger models (e.g., 13B parameters) in standard captioning metrics, including BLEU and METEOR, while maintaining substantially lower computational costs.
 
\end{abstract}



\maketitle

\begin{figure*}[!htbp]
  \centering
  \includegraphics[width=1\linewidth]{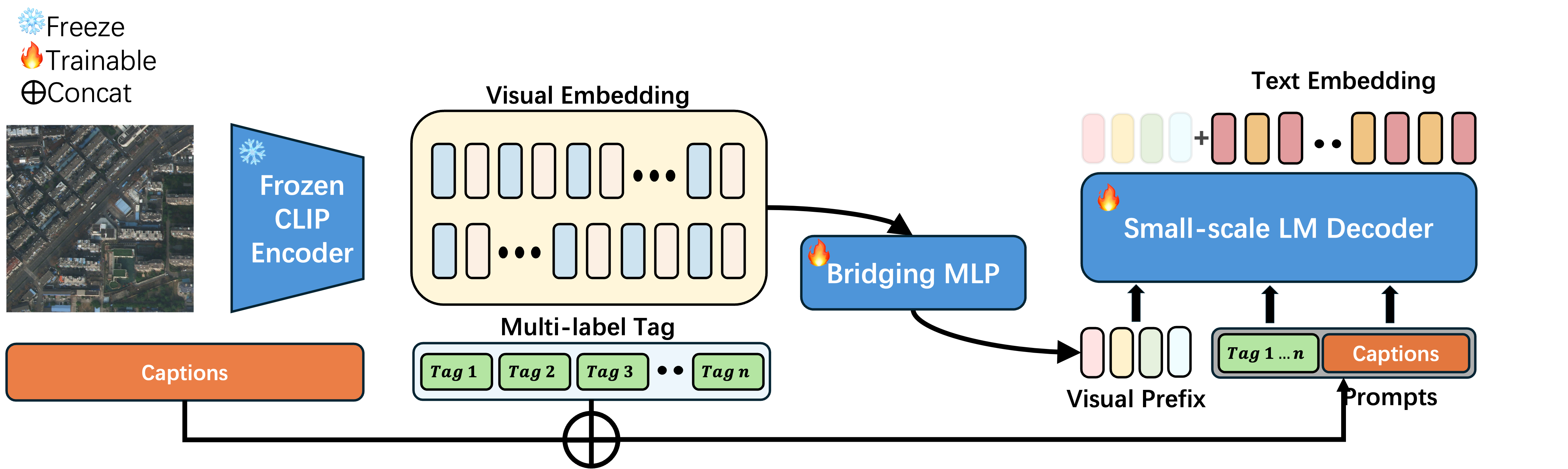}

  \caption{\textbf{Overall AeroLite pipeline for remote sensing captioning.}
  A frozen CLIP encoder (left) extracts a global image embedding (in orange) and predicts multi-label tags (in green). 
  The numeric embedding is passed through a bridging MLP to generate a sequence of visual tokens (shown in pink/white), 
  while the predicted tags are incorporated into a text prompt to form an instruction for the small-scale language model (right). 
  Through LoRA-based or prefix-only fine-tuning, the language model fuses visual tokens and tag-text tokens in a single 
  self-attention context, enabling high-quality captions with minimal computational cost.}

  \label{fig:aerolite_pipeline}
\end{figure*}

\section{introduction}
Remote sensing imagery is indispensable in areas such as environmental monitoring, urban planning, agricultural management, and disaster response \cite{intro}. Unlike standard photographs, aerial images often cover expansive regions, feature complex multi-object layouts, and demand specialized semantics—for instance, runways, oil tanks, or unique farmland patterns \cite{intro2}. However, producing high-quality annotations for these images remains prohibitively expensive, largely due to the need for expert-level interpretation \cite{intro3,intro4}. Consequently, \emph{automated remote sensing image captioning}—translating crucial visual details into natural language—has become a pressing goal for enabling quick, reliable insights for both decision-makers and automated systems.

While early efforts based on CNN-RNN pipelines (e.g., Show \& Tell \cite{CNN-RNN}) established a basic framework for image captioning, they were not well-suited to high-resolution aerial imagery \cite{RSICD-CNN-RNN,RS-CNN-LSTM}. Transformer-based models and attention mechanisms have improved alignment between vision and language, yet they typically require extensive domain-specific pretraining to grasp aerial semantics \cite{RS-Transformer-Caption, RS-Transformer-Caption2}. More recently, large language models (LLMs) such as GPT \cite{GPT-4}, OPT \cite{OPT}, and LLaMA \cite{llama} have inspired hybrid approaches that pair frozen vision encoders (e.g., CLIP \cite{clip}) with LLMs via prompting or prefix embeddings \cite{llava, blip-1, blip-2, flamingo}. Although effective for general-purpose imagery, these approaches encounter two critical hurdles in remote sensing:

\begin{enumerate} 
\item \textbf{Domain Gap:} Standard pretrained models seldom capture specialized aerial perspectives or terminology, leading to inaccuracies when directly applied to satellite imagery \cite{GeoChat, RS-Clip, RS-LLaVA}. 
\item \textbf{Resource Constraints:} Models ranging into billions of parameters demand substantial hardware. Even a 7B model (e.g., LLaMA-7B) often exceeds 8GB of half-precision memory, restricting their viability for on-platform uses such as drones or satellites \cite{7B-limitation, edge-limitation}. 
\end{enumerate}

In response, some researchers have proposed simpler “tag → LLM” pipelines, where multi-label classifiers produce semantic tags (e.g., “runway,” “forest”) to prompt mid-sized LLM \cite{tag2text}. Although lightweight and generally fluent, such pipelines overlook in-depth visual relationships by focusing solely on tags. On the other hand, large-scale vision–language frameworks like BLIP-2 \cite{li2023blip} and miniGPT \cite{chen2023minigptv2largelanguagemodel} offer robust multimodal grounding yet carry steep demands in training and inference. Taken together, these constraints underscore the need for an approach that employs explicit tag guidance \emph{and} direct visual grounding while staying within practical computational limits \cite{zheng2024instruction}.

In this paper, we introduce \textbf{AeroLite}, a tag-guided, LLM-agnostic framework designed for remote sensing image captioning. Our method fuses a CLIP-based multi-label classifier, a lightweight MLP bridging module, and a smaller language model (on the order of 1--3B parameters) to yield accurate, context-rich captions at manageable computational cost \cite{niraula2024multi}. Concretely, we start by using GPT-4\textsubscript{o} to generate large-scale pseudo-captions from remote sensing segmentation datasets, then extract high-confidence semantic tags through NLP to train a multi-label CLIP classifier. Next, we map CLIP embeddings into the LLM’s space via a compact MLP—avoiding heavy modifications to either component—and perform a two-stage LoRA \cite{lora} fine-tuning procedure: first acquiring remote sensing semantics through pseudo-labeled data, then refining caption style and domain alignment on smaller, real-world datasets.

Our key contributions include: 
\begin{itemize} 
\item \textbf{Lightweight, LLM-Agnostic Architecture:} We introduce a simple MLP bridging module adaptable to various language models (1–3B parameters or larger), significantly reducing computational overhead while preserving flexibility for user-preferred LLM backbones. 
\item \textbf{Tag-Guided Visual Alignment:} AeroLite combines explicit semantic tags for interpretability with direct CLIP visual features for detailed visual grounding, overcoming limitations inherent in purely tag-based pipelines. 
\item \textbf{Two-Stage LoRA Fine-Tuning:} We adopt a LoRA-based approach that first leverages pseudo-captioned data for semantic grounding, then fine-tunes for captioning style and domain-specific alignment using smaller real-world caption sets. 
\item \textbf{State-of-the-Art Performance:} AeroLite surpasses larger-scale baseline models (7B–13B parameters) on benchmark datasets (e.g. UCM, Sydney), providing enhanced explainability and controllability through tag-level insights crucial for real-world remote sensing applications. 
\end{itemize}

\begin{figure*}[htbp]
  \centering
  \includegraphics[width=0.95\linewidth]{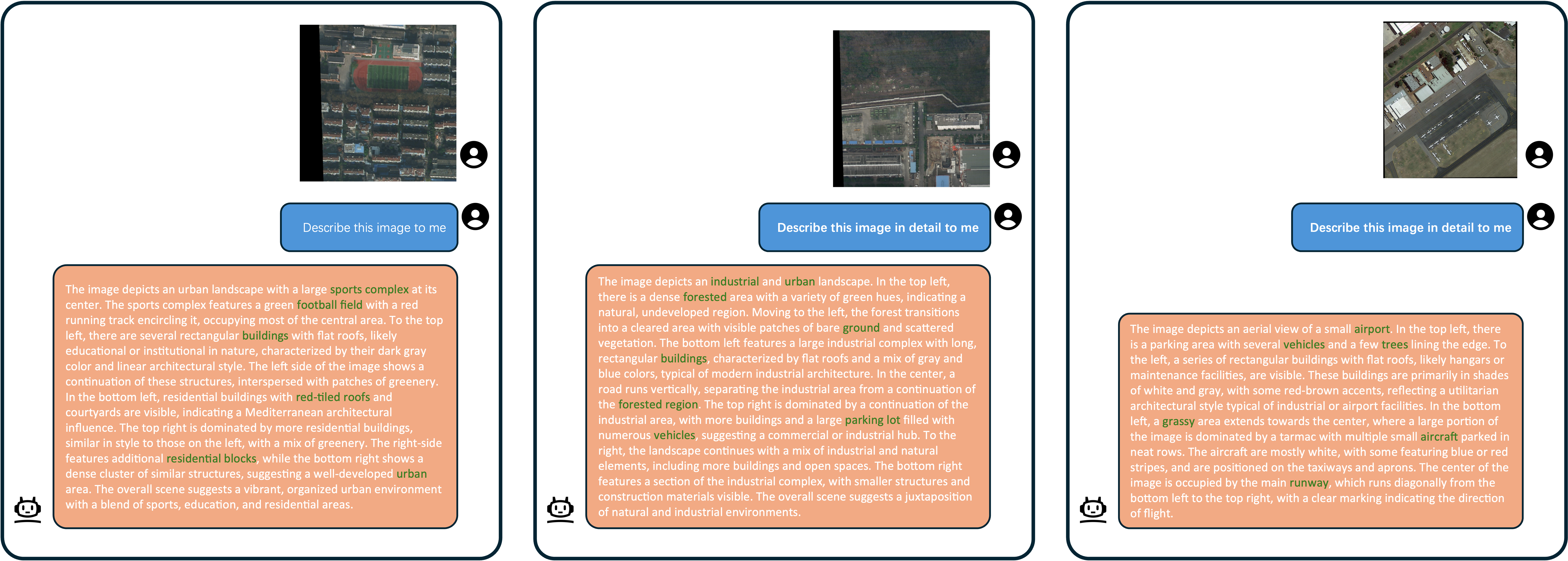}
  \caption{\textbf{AeroLite Inference on Small-Scale LM:} Example predictions on diverse aerial scenes, highlighting how explicit tags (in green) guide the language model to focus on specialized remote sensing semantics, 
  such as \emph{“industrial zone”} or \emph{“runway.”} }
  \label{fig:aerolite_taggedsamples}
\end{figure*}

\section{Related Work}
\label{sec:relatedwork}
In recent years, image captioning has seen substantial advances in both natural and remote sensing domains. Due to the complexity of aerial scenes (e.g., diverse land cover types, varying scales of objects, and data scarcity), multiple technical approaches have been proposed. These can be broadly categorized into four phases or methods: \textbf{(1) CNN+RNN Methods}, \textbf{(2) Transformer/Attention Methods}, \textbf{(3) LLM-Based Captioning with Image-Derived Tags}, and \textbf{(4) Pretrained Vision-Language Models}.

\subsection{CNN + RNN Captioning Models}
Early image captioning pipelines often adopted a classic\\ 'encoder-decoder' design, where a convolutional network (CNN) encodes the image into a feature representation, and a recurrent network (RNN) generates a textual description word by word. Representative works include Show \& Tell \cite{vinyals2015show} and Show, Attend and Tell \cite{xu2015show}, which pioneered this approach for natural images. In the remote sensing domain, Shi and Zou \cite{shi2017can} proposed one of the first CNN-LSTM frameworks specifically for aerial imagery.  
These methods typically train on datasets such as MS COCO, Flickr30k (for natural images), and smaller remote-sensing sets like Sydney/UCM Captions or the larger RSICD (with around 10k images) \cite{lu2017exploring}.  
One hallmark of this approach is its relatively straightforward architecture, which leverages pretrained CNNs for feature extraction and incorporates early attention mechanisms in the RNN decoder to better highlight key objects. However, when confronted with lengthy descriptions or complex scenes, RNN-based models can lose fine-grained details over long sequences. Coupled with the limited availability of remote sensing data, such constraints often lead to overfitting and diminished generalizability across diverse aerial environments.

\subsection{Transformer and Attention-Based Models} Subsequent research introduced stronger attention mechanisms and fully Transformer-based architectures, exemplified by the “Bottom-Up and Top-Down” attention model \cite{anderson2018bottom}, which employs Faster R-CNN to extract region-level features and uses an attention-based LSTM to focus on these regions during text generation. This paradigm achieved substantial performance gains on datasets like MS COCO, largely thanks to its ability to attend to multiple salient objects or areas in parallel. In the meantime, the introduction of the Transformer \cite{vaswani2017attention} enabled parallelized modeling of long-range dependencies through multi-head self- and cross-attention, entirely replacing RNNs in some architectures. A notable example is RSTNet \cite{zhang2021rstnet}, which leverages adaptive attention to balance visual features with linguistic context and reports impressive results on COCO. In the remote sensing domain, Gajbhiye and Nandedkar \cite{gajbhiye2022generating} integrated a CNN-based encoder with a Transformer decoder augmented by spatial and channel-wise attention, effectively handling varied landscapes and small objects. Although these approaches generally produce more detailed and coherent captions, they often come with considerably larger parameter counts, thus requiring more data during training. In remote sensing settings, limited annotated data can lead to overfitting if not mitigated through measures such as regularization or domain-oriented pretraining. Region detection can also present a bottleneck, as overlooked targets may never surface in the generated captions.

\subsection{LLM-Based Captioning with Image-Derived Tags} A more recent trend leverages Large Language Models (LLMs) by feeding them image-derived tags or attributes instead of raw visual features. Under this two-stage paradigm, a dedicated vision model first extracts key information, such as objects, attributes, or land-cover categories—before passing the resulting textual tokens to a pretrained LLM (e.g. ChatGPT). CapText \cite{ghosh2023captext} showed that, given a collection of object labels or brief descriptors, an LLM can produce fluent captions without directly accessing the original pixels. In the remote sensing context, practitioners can apply a multi-label classifier or object detector to identify elements like “forest,” “river,” or “buildings,” then optionally fine-tune the LLM to boost domain alignment. This modular strategy capitalizes on the LLM’s robust language modeling and extensive real-world knowledge while minimizing additional data requirements, but any omission or error introduced by the tagger directly manifests in the final caption \cite{dong2024benchmarking, sarto2025image}. Moreover, because the LLM lacks direct visual grounding, ambiguities in the prompt or mistaken labels can lead to “hallucinated” details or incomplete scene descriptions \cite{bai2024hallucination}.

\begin{figure*}[t]
\centering
\fcolorbox{black}{gray!20}{
\begin{minipage}{1\linewidth}
\vspace{0.5em}

\begin{minipage}[t]{0.58\linewidth}
\textbf{Prompt Example}\\[0.5em]
\small\ttfamily
You are a professional geography scene description expert. \\
Given a remote sensing image and its corresponding polygon-based annotations (including approximate coordinates and categories), provide a single-sentence description focusing on these key points:\\

1) Use relative positional terms (e.g., <left side>, <right side>, <top>, <bottom>, <center>) to describe the locations of main features.

2) If a category occupies a significantly large portion of the image, emphasize it with phrases such as <most of> or <large portion of>.

3) Remain objective and concise, avoiding unnecessary adjectives; emphasize relative positioning and approximate area coverage.\\[0.5em]

Verify your description by cross-checking:\\
- The visual content of the image,\\
- The provided annotation data (approximate coordinates, categories).\\[0.5em]

Your response must strictly adhere to these specifications.
\end{minipage}
\hfill
\begin{minipage}[t]{0.38\linewidth}
\centering
\textbf{Image:}\\[0.5em]
\includegraphics[width=0.7\linewidth]{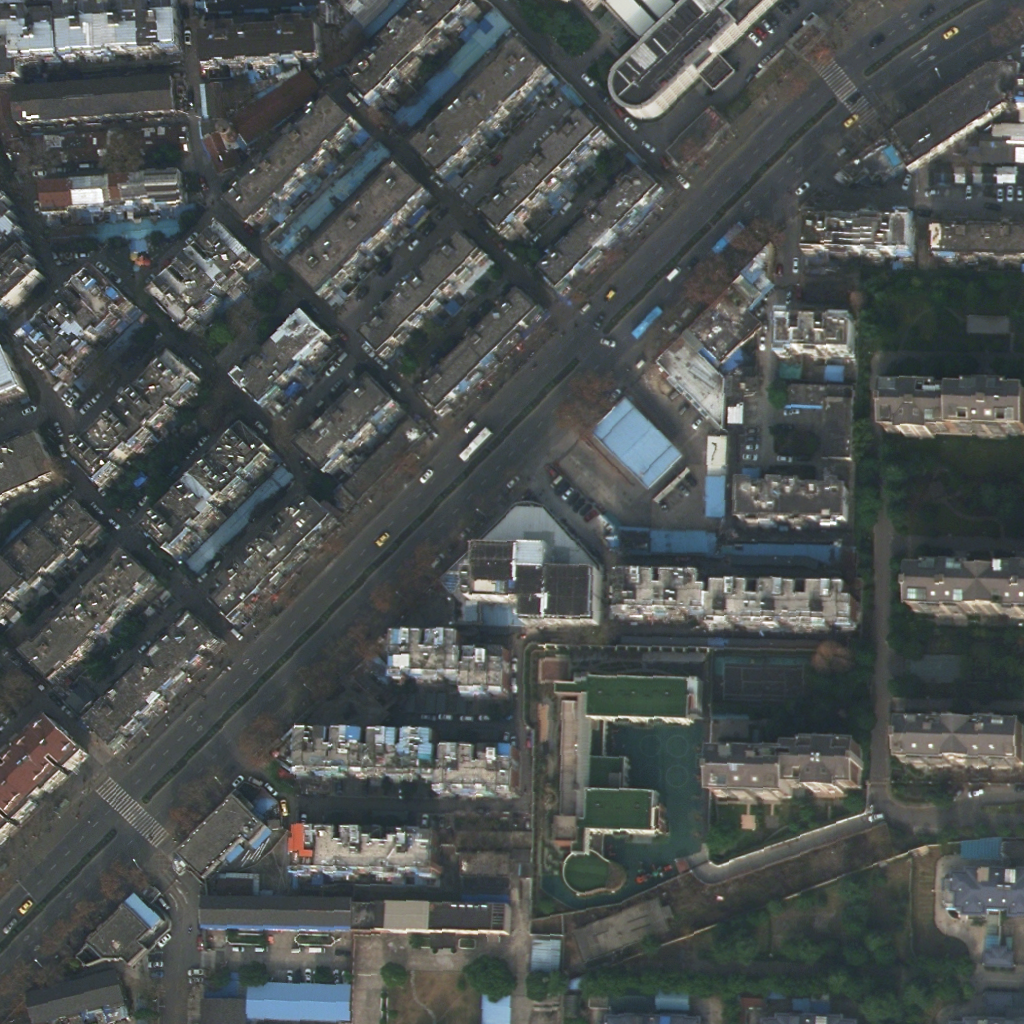}\\[1em]

\raggedright
\textbf{Semantic Segmentation Labels:}\\[0.5em]
building [0.6, 0.5, ...],building [0.3, 0.1, ...], road [0.2, 0.8, ...],
trees [0.7, 0.2, ...], object: [...] \textless omitted\textgreater
\end{minipage}

\vspace{1em}
\begin{minipage}{\linewidth}
\noindent\rule{\linewidth}{0.4pt}
\textbf{Generated Description:}\\[0.5em]
Most of the right and central parts of the image are occupied by densely packed buildings, a wide road runs vertically along the middle-left portion, and a playground area with green fields is located near\textless omitted\textgreater
\end{minipage}

\vspace{0.5em}
\end{minipage}
}
\caption{Illustration of the prompt layout, remote sensing image, semantic segmentation labels, and the generated description.}
\label{fig:remote-sensing-prompt}
\end{figure*}

\subsection{Pretrained Vision-Language Models (VLMs)} The current state-of-the-art in image captioning is now dominated by large-scale pretrained Vision-Language Models (VLMs) such as Flamingo \cite{alayrac2022flamingo}, BLIP-2 \cite{li2023blip}, and InstructBLIP \cite{instructblip}, which typically couple a frozen CNN or ViT backbone with a frozen LLM (GPT-like or T5) through a smaller learnable module (e.g., a Perceiver Resampler or Q-Former). Flamingo, for instance, leverages its Perceiver Resampler to convert feature maps into a set of visual tokens for a 70B Chinchilla LLM, whereas BLIP-2 employs a Q-Former to align visual features with the LLM’s latent space. After training on vast multimodal corpora, these models can tackle tasks such as zero-shot captioning and VQA with minimal additional effort. In the remote sensing arena, RS-CapRet \cite{RS-CapRet} adapts this approach by combining a CLIP-based encoder, a frozen LLM, and a compact adapter to achieve state-of-the-art results on RSICD. Although these models excel at producing context-rich, highly detailed descriptions and can transfer to new tasks with comparatively little fine-tuning, their massive scale poses challenges for real-time or on-device deployment. They can also produce “hallucinated” content when the LLM invokes prior knowledge not directly obtained from the image, making domain adaptation crucial for specialized datasets such as satellite imagery \cite{li2023evaluating}. The computational cost and relative opacity of their multi-modal internals further underscore the complexity of deploying them in resource-constrained environments or scenarios requiring high interpretability \cite{sun2024review}.

\section{Data Generation}
\label{sec:data-gen}
Remote sensing has seen remarkable progress in segmentation datasets over the past few years. Popular benchmarks such as \textbf{RSSCN7}~\cite{zou2015deep}, \textbf{DLRSD}~\cite{li2023msanet}, \textbf{iSAID}~\cite{waqas2019isaid}, \textbf{LoveDA}~\cite{wang2021loveda}, and \textbf{WHU}~\cite{ji2018fully} collect a vast array of high-resolution aerial or satellite images, accompanied by polygon-based annotations of diverse geographic objects (e.g., buildings, farmland, rivers). These annotations precisely outline object boundaries, enabling robust supervised learning in various scene understanding tasks~\cite{llava,RS-LLaVA}. However, as illustrated in Figure~\ref{fig:remote-sensing-prompt}, most of these datasets only provide polygon masks and categories, lacking \emph{descriptive captions} about spatial layouts or relationships. This limitation poses a significant challenge for multimodal remote sensing analysis.

Despite recent efforts such as \textbf{RSICD}, which strives to introduce more interpretable textual annotations, \emph{truly} multimodal datasets (pairing polygons and captions) remain scarce. Given the cost and difficulty of manually writing high-quality captions for large-scale aerial imagery, we propose a two-step pipeline: \textbf{(1)}~Automatically generate an initial set of textual descriptions using GPT; and \textbf{(2)}~Extract essential \emph{semantic tags} to guide visual representation learning.

\subsection{GPT-assisted Caption Generation}
\label{subsec:auto-caption}
Formally, for each remote sensing image \(X_v\) with polygon annotations \(X_p\), we build a structured prompt \(P(X_p)\) that lists both categorical labels (e.g.\ \emph{building}, \emph{river}) and approximate coordinates in a concise, machine-readable format. An example prompt is shown in Figure~\ref{fig:remote-sensing-prompt}, which includes relative positions like \texttt{[0.6, 0.5, ... ,0.3]}. We then use \textbf{GPT-4o} to generate short but informative pseudo-captions:
\[
C = \mathrm{GPT\text{-}4o}\bigl(P(X_p)\bigr).
\]
Specifically, GPT-4o is instructed to \emph{(i)}~describe major objects or regions with terms such as \emph{“most of”}, \emph{(ii)}~use relative positioning (e.g.\ \emph{top-right}, \emph{bottom-left}), and \emph{(iii)}~avoid unnecessary adjectives. This automated approach \textbf{significantly reduces} annotation costs while retaining spatially accurate textual information about each polygon-annotated image. Empirical tests indicate that these GPT-based captions preserve key semantic and positional details~\cite{VRSBench,llava}.

\begin{figure}[!ht]
  \centering
  \includegraphics[width=0.48\textwidth]{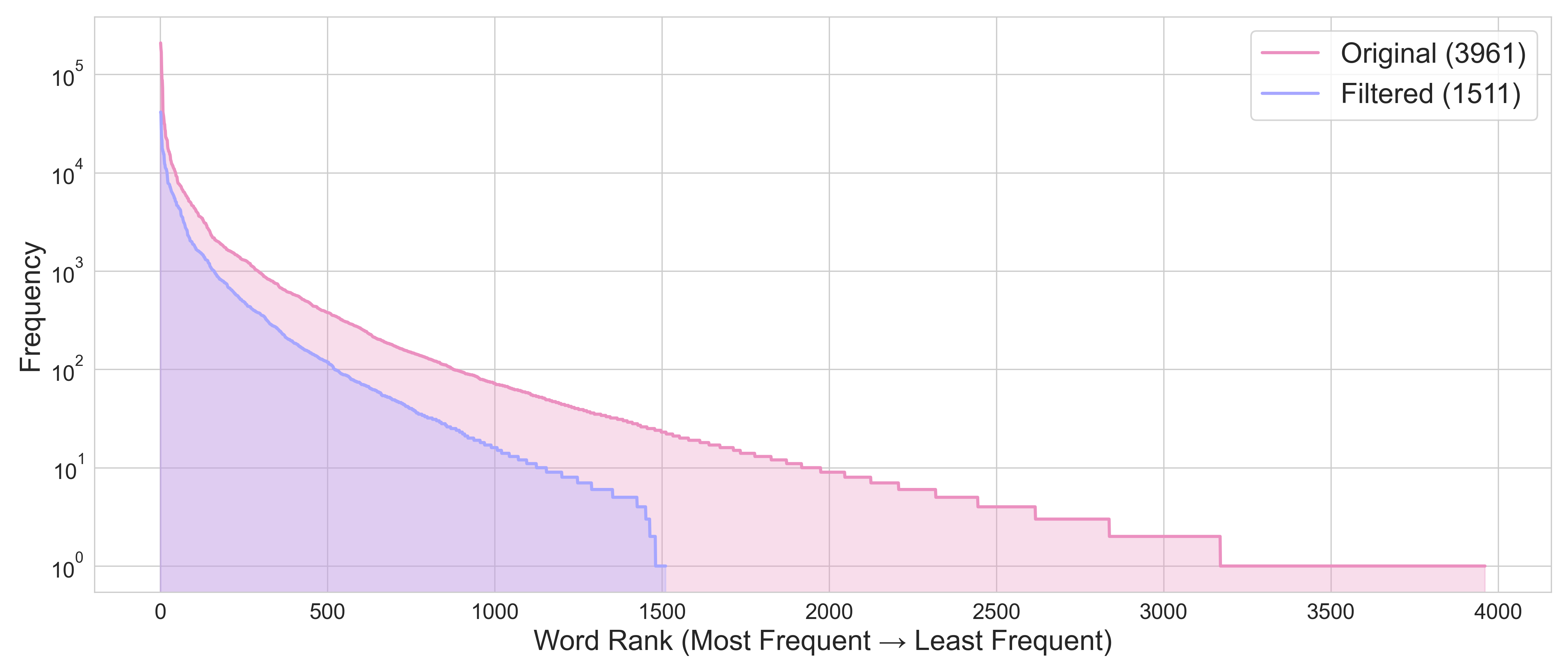}
  \caption{Comparison of the original (pink) vs.\ filtered (blue) vocabulary distributions on a log scale. The filtering process discards rare and noisy terms, resulting in a more compact yet expressive vocabulary.}
  \label{fig:freq_comparison}
\end{figure}

\paragraph{Vocabulary Filtering.} Because GPT-4\textsubscript{o} may introduce rare or noisy tokens, we first analyze the global token frequency across all generated captions and prune those that appear below a specified threshold. As visualized in Figure~\ref{fig:freq_comparison}, this process (blue bars) significantly reduces the vocabulary size compared to the unfiltered set (pink bars), removing redundant or idiosyncratic words while preserving essential terms for remote sensing descriptions. The resulting corpus covers \textbf{12{,}473} images with captions averaging \textbf{181.94 words} and \textbf{9.21 sentences}, yielding \textbf{2507} unique words (including \textbf{1557} frequent noun tokens such as \emph{“Vessel,” “Runway,”} and \emph{“Vehicle”}).

\paragraph{Semantic Tag Extraction for Visual Training.} Following vocabulary filtering, we derive a concise set of highly relevant words from each caption to serve as \emph{semantic tags}. As shown on Algorithm 1, we apply named entity recognition (NER) and part-of-speech tagging to isolate key geographical or scene-related nouns and modifiers, which collectively form a multi-label Tag set \(\mathbf{T}\) for each image. By retaining only terms above a minimum frequency (and discarding low-value tokens), we ensure that the extracted tags accurately capture high-level domain cues (e.g., \emph{“runway”}, \emph{“oil tank”}, \emph{“grain field”}) without overwhelming the model with extraneous language. These discrete tags are then used in our downstream multi-label classification approach, strengthening the visual encoder’s ability to interpret specialized aerial semantics and bridging the gap to high-quality caption generation.

\begin{algorithm}[H]
\caption{Semantic Tag Creation (Extracting Nouns and Adjectives)}
\begin{algorithmic}[1]
\Require Generated caption set \(C\)
\Ensure Semantic tag set \(T\)
\State \(T \gets \emptyset\)
\For{each caption \(c \in C\)}
    \State Perform tokenization on caption \(c\)
    \State Apply NER to identify geographic <entities>
    \State Extract identified "Nouns" and "Adjectives"
    \State Update tag set: \(T \gets T \cup \{\text{extracted words}\}\)
\EndFor
\State \Return \(T\)
\label{Tag Creation}
\end{algorithmic}
\end{algorithm}

\section{Methodology}
\subsection{AeroLite Model Architecture}
\label{sec:aerolite-arch}

In this section, we introduce our \textbf{AeroLite} framework for remote sensing image captioning, which comprises three key components:
\begin{enumerate}
    \item A \textbf{CLIP-based multi-label visual encoder} that extracts high-level image features and semantic tags,
    \item A \textbf{lightweight bridging MLP} that aligns these CLIP-derived features to a small-scale LLM’s embedding space,
    \item A \textbf{small-scale language model}, fine-tuned via \textbf{LoRA} for domain-specific adaptation.
\end{enumerate}
As depicted in Figure~\ref{fig:aerolite_pipeline}, AeroLite first employs a frozen CLIP encoder to generate image embeddings and semantic tags, then fuses them in the LLM through the bridging MLP and tag-based prompts. The following sections detail these design choices, explaining how they balance computational efficiency with high-quality caption generation across large-scale remote sensing scenarios.

As we detail below, these choices strike a balance between computational efficiency and caption generation quality, making them well-suited for large-scale remote sensing imagery.

\subsection{CLIP-based Multi-label Visual Encoder}
Remote sensing images often contain multiple categories (e.g., buildings, farmland, vehicles) distributed across large spatial areas. Rather than classifying a single label, we adopt a multi-label classification strategy. Specifically, we select CLIP for its robust zero-shot generalization properties~\cite{clip, liang2024survey}. Although originally trained on natural images, CLIP still adapts well to aerial imagery under minimal supervision~\cite{liu2024remoteclip, li2023rs}.

We \emph{freeze} the CLIP backbone and attach a linear classifier for multi-label tagging:
\begin{equation}
\label{eq:tagging_head}
\mathbf{p} 
\,=\,
\sigma\!\Bigl(\mathbf{W}_{\mathrm{tag}}\,\mathbf{v} \,+\, \mathbf{b}_{\mathrm{tag}}\Bigr),
\end{equation}
where \(\mathbf{v}\) denotes the CLIP-encoded feature, and \(\mathbf{p}\in[0,1]^{K}\) is a probability vector over \(K\) tags. We use widely accepted probability \(\tau = 0.5\) \cite{lin2024tagclip} to infer the presence of each label. This yields both high-quality visual embeddings \(\mathbf{v}\) and explicit semantic tags \(\mathbf{T}_{\mathrm{pred}}\) for the next stage.

\paragraph{Multi-label Classification Loss.}
Given a ground-truth binary vector \(\mathbf{y}\in\{0,1\}^K\), we adopt a multi-label binary cross-entropy loss:
\begin{equation}
\label{eq:ml_tag_loss}
\mathcal{L}_{\mathrm{tag}}
=
-\sum_{k=1}^{K}
\Bigl[
\,y_k\log(p_k)\;+\;(1-y_k)\log(1-p_k)\Bigr].
\end{equation}
We only update \(\mathbf{W}_{\mathrm{tag}}, \mathbf{b}_{\mathrm{tag}}\), keeping the CLIP backbone frozen.

\subsection{Lightweight Visual-Language Bridging MLP}
In visual language fusion tasks, a key question is: how can visual features be effectively incorporated into the input space of the language model? Although more complex cross-modal attention structures (such as Q-Former) can provide deeper interactions, research has shown (LLaVA\cite{llava} and MiniGPT-4\cite{zhu2023minigpt4enhancingvisionlanguageunderstanding}) that a simple MLP can achieve efficient alignment as long as the features provided by the visual encoder are strong enough.

We next map \(\mathbf{v}\) (and optionally the predicted tags \(\mathbf{T}_{\mathrm{pred}}\)) into the LLM's token embedding space. A simple MLP suffices when robust features are provided by CLIP. Concretely,
\begin{equation}
\label{eq:mlp_bridge}
\begin{aligned}
\mathbf{h} 
&= 
\operatorname{ReLU}(\mathbf{W}_1\,\mathbf{v} + \mathbf{b}_1),\quad
\mathbf{z}
=
\mathbf{W}_2\,\mathbf{h}
+ 
\mathbf{b}_2,
\end{aligned}
\end{equation}
yielding a \(\mathbf{z}\in\mathbb{R}^{d_z}\). In practice, we replicate or extend \(\mathbf{z}\) into multiple consecutive “visual prefix tokens” and prepend them to the LLM input. Since these tokens reside in the same space as regular text embeddings, the LLM can seamlessly incorporate image context without architectural modification.

\begin{algorithm}[H]
\caption{AeroLite: Concise Training Pipeline}
\label{alg:aerolite_concise}
\begin{algorithmic}[1]
\Require
  \(\mathcal{D} = \{\!(\mathbf{X}_i, \mathbf{Y}_i, \mathbf{T}_i)\!\}\): dataset (images \(\mathbf{X}_i\), captions \(\mathbf{Y}_i\), optional tags \(\mathbf{T}_i\)) \\
  Frozen CLIP encoder \(\mathcal{E}_{\mathrm{CLIP}}\), \\
  Tag head \((\mathbf{W}_{\mathrm{tag}}, \mathbf{b}_{\mathrm{tag}})\), \\
  Bridging MLP \(\theta_{\mathrm{MLP}}\), \\
  LLM w/ LoRA factors \(\{\mathbf{A}_\ell,\mathbf{B}_\ell\}\), optional unfreeze top \(N\) base weights \(\{\mathbf{W}_\ell^{(\mathrm{base})}\}\), \\
  Hyperparams: \(\tau, \alpha, \eta\), epochs \(E\).
\Ensure Updated model parameters
\For{epoch = 1 \textbf{to} E}
  \For{each mini-batch \(\{\!(\mathbf{X}_j,\mathbf{Y}_j,\mathbf{T}_j)\!\}\)}
    \State \(\mathbf{v}_j \gets \mathcal{E}_{\mathrm{CLIP}}(\mathbf{X}_j)\) \Comment{CLIP is frozen}
    \State \(\mathbf{p}_j \gets \sigma(\mathbf{W}_{\mathrm{tag}}\mathbf{v}_j + \mathbf{b}_{\mathrm{tag}})\)
    \State optional \(\mathcal{L}_{\mathrm{tag}}\) via BCE if \(\mathbf{T}_j\) exists
    \State \(\mathbf{z}_j \gets \theta_{\mathrm{MLP}}(\mathbf{v}_j, \{\!k\mid p_{j,k}\!\ge\!\tau\!\})\)
    \State \(\hat{\mathbf{Y}}_j \gets \mathcal{F}_{\mathrm{LLM}}(\mathbf{z}_j \mid \{\mathbf{A}_\ell,\mathbf{B}_\ell\}, \{\mathbf{W}_\ell^{(\mathrm{base})}\})\)
    \State \(\mathcal{L}_{\mathrm{cap}}\) compares \(\hat{\mathbf{Y}}_j\) w/ \(\mathbf{Y}_j\)
    \State \(\mathcal{L}_{\mathrm{total}} \gets \mathcal{L}_{\mathrm{cap}} + \alpha\,\mathcal{L}_{\mathrm{tag}}\)
    \State \textbf{update} 
      $\{\mathbf{W}_{\mathrm{tag}},\mathbf{b}_{\mathrm{tag}},\theta_{\mathrm{MLP}},\mathbf{A}_\ell,\mathbf{B}_\ell,\mathbf{W}_\ell^{(\mathrm{base})}\}$
      \textbf{w.r.t.} $\mathcal{L}_{\mathrm{total}}$

  \EndFor
\EndFor
\end{algorithmic}
\end{algorithm}

\subsection{Small-scale Language Model Fine-tuned via LoRA}
\label{subsec:lora}

Recent studies show that \emph{small-scale} LLMs (1-3B parameters) can achieve near or even surpass commercial LLMs (e.g., GPT-4o, Gemini\cite{team2023gemini}) on certain tasks when guided properly\cite{dey2023btlm}. In our framework, we adopt a smaller LLM (e.g., LLaMA-3B or Gemma-2B) and further reduce training overhead via Low-Rank Adaptation (\textbf{LoRA})~\cite{lora}. 

\paragraph{Key Formulae in the Text.}
LoRA inserts trainable low-rank matrices \(\mathbf{A}, \mathbf{B}\) into specific linear layers of the LLM, while the original large weights \(\mathbf{W}\) remain \emph{frozen}. Formally, for a hidden vector \(\mathbf{H}\), the re-parameterization is:
\begin{equation}
\label{eq:lora_eq}
\mathbf{H}' 
\;=\;
\mathbf{H} 
\;+\;
\mathbf{A}\Bigl(\mathbf{B}\,\mathbf{H}\Bigr),
\end{equation}
which drastically limits the number of updated parameters. Additionally, if resources allow, we may \emph{partially unfreeze} the top $N$ layers of the LLM. In these layers, the forward pass is:
\begin{equation}
\label{eq:partial_unfreeze_eq}
\mathbf{W}_\ell^\prime
\;=\;
\mathbf{W}_\ell^{(\mathrm{base})}
\;+\;
\mathbf{A}_\ell\mathbf{B}_\ell,
\quad
\mathbf{H}_\ell^\prime
\;=\;
\mathbf{W}_\ell^\prime\,\mathbf{H}_\ell,
\end{equation}
where both \(\mathbf{W}_\ell^{(\mathrm{base})}\) (the base weight) and \(\{\mathbf{A}_\ell,\mathbf{B}_\ell\}\) (LoRA factors) are updated by backpropagation. For the remaining layers, we keep \(\mathbf{W}_\ell^{(\mathrm{base})}\) frozen, allowing partial adaptation without overburdening computation.

We summarize the complete training procedure in Algorithm~\ref{alg:aerolite_concise}. In essence, the bridging MLP (\(\theta_{\mathrm{MLP}}\)) and multi-label head \((\mathbf{W}_{\mathrm{tag}}, \mathbf{b}_{\mathrm{tag}})\) are always trainable, while LoRA operates on selected LLM layers. Optionally, we unfreeze the top $N$ LLM layers for additional fine-tuning capacity. The total loss function comprises the language modeling term plus the optional multi-label classification term (Eq.~\eqref{eq:ml_tag_loss}).




\begin{table*}[t]
\centering
\caption{Performance comparison between CLIP backbones on our aerial multi-label dataset.}
\label{tab:clip_compare}
\small
\begin{tabular}{lcccccccccc}
\toprule
\textbf{Backbone} & \textbf{Params} & \textbf{GPU Mem} & \textbf{P@10} & \textbf{R@10} & \textbf{F1@10} & \textbf{mAP@10} & \textbf{R@1} & \textbf{R@5} & \textbf{R@10} & \textbf{Epoch Time} \\
\midrule
\textbf{ViT-B/32} & 151.9M & $\sim$3.4GB & 33.11\% & 73.20\% & 42.59\% & 59.29\% & 82.98\% & 96.18\% & 98.81\% & \textbf{15:54} \\
\textbf{ViT-L/14} & 428.5M & $\sim$6.5GB & \textbf{36.36\%} & \textbf{78.31\%} & \textbf{46.28\%} & \textbf{67.11\%} & \textbf{87.51\%} & \textbf{98.45\%} & \textbf{99.38\%} & 20:54 \\
\bottomrule
\end{tabular}
\end{table*}

\begin{table*}[t]
\centering
\caption{Comparison of representative CNN methods, popular VLM methods, and our FullTune approach on \textbf{UCM} and \textbf{Sydney} Captions datasets. 
Metrics include BLEU-1, BLEU-4, METEOR, and ROUGE-L.}
\label{tab:cnn_vlm_fulltune}
\small
\begin{tabular}{lllcccccccc}
\toprule
\textbf{Method} & \textbf{Vis Encoder} & \textbf{Text Model} 
& \multicolumn{4}{c}{\textbf{UCM Captions}} 
& \multicolumn{4}{c}{\textbf{Sydney Captions}} \\
\cmidrule(lr){4-7} \cmidrule(lr){8-11}
& & 
& \textbf{B1} & \textbf{B4} & \textbf{M} & \textbf{R} 
& \textbf{B1} & \textbf{B4} & \textbf{M} & \textbf{R} \\
\midrule
VLAD+RNN \cite{Lu_2018}
    & VGG19 & RNN
    & 63.11 & 42.09 & 29.71 & 58.78
    & 56.58 & 32.79 & 26.72 & 52.71 \\

Hard-attention \cite{pmlr-v37-xuc15}
    & VGG16 & LSTM
    & 81.57 & 61.82 & 42.63 & 76.98
    & 75.91 & 52.58 & 38.98 & 71.89 \\

SVM-D BOW \cite{9521989}
    & VGG16 & SVM
    & 76.35 & 51.95 & 36.54 & 68.01
    & 77.87 & 53.05 & 37.97 & 69.92 \\
\midrule
    

AMHT \cite{amht}
    & CLIP & GPT2
    & 84.90 & 69.90 & 40.90 & 46.70
    & 87.80 & 70.10 & 39.00 & 42.80 \\

CRSR \cite{wang2024crossmodalretrievalsystematicreview}
    & CLIP ViT-L/14 & LLaMA2-7B
    & 90.60 & 76.81 & 49.56 & 85.86
    & 79.94 & 66.02 & 41.50 & 74.88 \\

RS-CapRet \cite{RS-CapRet}
    & CLIP-Cap-4 & LLaMA2-7B
    & 84.30 & 67.00 & 47.20 & 81.70
    & 78.70 & 56.40 & 38.80 & 70.70 \\

SkyEyeGPT-7B \cite{zhan2025skyeyegpt}
    & EVA-G & LLaMA2-7B
    & 90.71 & 78.41 & 46.24 & 79.49
    & 91.85 & 77.40 & 46.62 & 77.74 \\

RSGPT-13B \cite{RSGPT13B}
    & EVA-G & Vicuna-13B
    & 86.12 & 65.74 & 42.21 & 78.34
    & 82.26 & 62.23 & 41.37 & 74.77 \\
\midrule
\textbf{AeroLite (Ours)} 
    & CLIP ViT-L/14 & LLaMA3.2-3B (FullTune)
    & \textbf{93.41} & \textbf{79.61} & \textbf{49.82} & \textbf{88.01} 
    & \textbf{91.89} & \textbf{75.88} & \textbf{47.53} & \textbf{83.66} \\
\bottomrule
\end{tabular}
\end{table*}

\section{Experiments}
\label{sec:experiments}

In this section, we describe how our \textbf{AeroLite} framework is set up, trained, and evaluated for remote sensing image captioning. We begin with implementation details (\S\ref{subsec:impl-setup}) and dataset construction (\S\ref{subsec:data-pseudo}), followed by our training strategy (\S\ref{subsec:train-strategy}). We then investigate the effect of different CLIP backbones (\S\ref{subsec:clip-compare}) on multi-label classification before moving to the full captioning experiments.

\subsection{Implementation Setup}
\label{subsec:impl-setup}
We implement AeroLite in PyTorch and conduct all experiments on a single NVIDIA RTX~4090 GPU with half-precision (fp16). 
For multi-label prediction, we freeze a \texttt{CLIP ViT-L/14} backbone, attaching a lightweight linear head that predicts 1,500 common tags, a number determined by frequency analysis on our pseudo-caption corpus. 
Training uses a batch size of 32, a learning rate of $1\times10^{-5}$, and a temperature of 0.07 for contrastive alignment, running for at most 50 epochs with early stopping if validation mAP stagnates for 5 epochs.

\subsection{Datasets for Pseudo-Caption Generation}
\label{subsec:data-pseudo}
We merge five remote sensing segmentation datasets—RSSCN7~\cite{zou2015deep}, DLRSD~\cite{li2023msanet}, iSAID~\cite{waqas2019isaid}, LoveDA~\cite{wang2021loveda}, WHU~\cite{ji2018fully}—to obtain a broad range of aerial environments (urban, rural, forest, water). 
Each dataset includes polygon-based labels, which we convert into about 12,000 automatically generated GPT captions (see \S\ref{subsec:auto-caption}). 
These “pseudo-captions” capture domain-specific semantics (e.g.\ farmland patterns, large industrial zones) and serve as a large training pool to embed specialized aerial knowledge into our system.

\subsection{Datasets for Evaluation}
We test our method on two smaller captions dataset:
\begin{itemize}
    \item \textbf{UCM-Captions}~\cite{9868765}, derived from UCMerced Land-Use\cite{10.1145/1869790.1869829}, containing 2,100 images of size $256\times256$ across 21 categories. Each image has five unique captions, giving a total of 10,500 descriptions.
    \item \textbf{Sydney Captions}~\cite{10.1145/1869790.1869829}, consisting of 613 images spanning seven land-use types. Each image is $500\times500$ pixels, each with five caption sentences, highlighting key features such as farmland or urban blocks.
\end{itemize}

\subsection{Training Strategy}
\label{subsec:train-strategy}
Our system first learns remote sensing semantics by training on the large pseudo-caption corpus. We then conduct \emph{instruction-based refinement} on the smaller UCM and Sydney sets to align the output style with standard captioning metrics. Throughout this two-stage process, we compare two main approaches to adapting the language model (1--3B parameters):
\begin{itemize}
    \item \textbf{VisualPrefix (MLP only)}—the language model backbone is entirely frozen, and only a small bridging MLP receives gradients.
    \item \textbf{Partial Unfreeze \& LoRA}—we unfreeze approximately 30\% of the top LLM layers and apply LoRA~\cite{lora} for parameter-efficient domain adaptation.
\end{itemize}

Unless otherwise stated, we use a beam size of 1 or top-$k$ sampling ($k=50$) for inference. 
Evaluation focuses on BLEU-1/4, METEOR, and ROUGE-L scores.

\subsection{CLIP Backbone Comparison}
\label{subsec:clip-compare}
Before proceeding to full captioning, we first compare two CLIP architectures—\textbf{ViT-B/32} (152M parameters) and \textbf{ViT-L/14} (428M parameters) for multi-label classification on our aggregated aerial dataset. Both models share the same training hyperparameters (batch size=32, learning rate=$1\times10^{-5}$). Table~\ref{tab:clip_compare} reports their respective metrics (F1@10, mAP@10), GPU usage, and per-epoch runtime.


As shown in Table~\ref{tab:clip_compare}, \textbf{ViT-L/14} consistently achieves higher F1 and mAP, raising F1 from 42.59\% to 46.28\% and mAP from 59.29\% to 67.11\%, albeit at the cost of more memory and longer epochs (20:54 vs.\ 15:54). Given these substantial gains, we adopt \textbf{ViT-L/14} as our default CLIP backbone in subsequent experiments, ensuring a stronger visual representation for remote sensing imagery.

\noindent
With the choice of ViT-L/14 established, we now proceed to evaluate \textbf{AeroLite} in its entirety on UCM and Sydney, benchmarking against prior CNN-based methods and large-scale VLMs, as well as analyzing the impact of partial unfreeze, LoRA, and explicit tagging.

\begin{table*}[!ht]
\centering
\caption{
Comparison of \textbf{VisualPrefix} (\emph{MLP Only}) vs. \textbf{Partial Unfreeze \& LoRA} 
under \textbf{AeroLite (Ours)} across different language models on the \textbf{UCM} and \textbf{Sydney} datasets. 
We report BLEU-1 (\textbf{B1}), BLEU-4 (\textbf{B4}), METEOR (\textbf{M}), and ROUGE-L (\textbf{R}). 
Here, \emph{MLP Only} indicates that the LLM backbone is \emph{frozen} 
while only a small bridging MLP is trained, 
whereas \textbf{Partial Unfreeze \& LoRA} indicates that 30\% of the top layers are partially unfrozen 
and trained via LoRA. 
The better row in each pair is highlighted, and better scores are \textbf{bolded}.
}
\label{tab:visualprefix_vs_partial_lora}
\small
\begin{tabular}{l l l l c c c c c c c c}
\toprule
\multicolumn{4}{c}{\textbf{AeroLite (Ours)}} 
& \multicolumn{4}{c}{\textbf{UCM Captions}} 
& \multicolumn{4}{c}{\textbf{Sydney Captions}} \\
\cmidrule(lr){1-4} \cmidrule(lr){5-8} \cmidrule(lr){9-12}
& \textbf{Language Model} & \textbf{\#Params} & \textbf{Tuning} 
& \textbf{B1} & \textbf{B4} & \textbf{M} & \textbf{R} 
& \textbf{B1} & \textbf{B4} & \textbf{M} & \textbf{R} \\
\midrule

& Phi-4 Mini Instruct & 1.8B & VisualPrefix 
  & 67.39 & 27.93 & 27.71 & 52.55
  & 81.57 & 55.63 & 43.84 & 71.64 \\

\rowcolor{gray!15}
& Phi-4 Mini Instruct & 1.8B & Partial Unfreeze \& LoRA     
  & \textbf{75.47} & \textbf{35.63} & \textbf{33.25} & \textbf{59.66}
  & \textbf{91.72} & \textbf{73.03} & \textbf{49.25} & \textbf{78.65} \\

& Gemma 2 & 2B & VisualPrefix 
  & 83.15 & 60.92 & 45.65 & 79.32 
  & 85.69 & 62.91 & 44.80 & 68.10 \\

\rowcolor{gray!15}
& Gemma 2 & 2B & Partial Unfreeze \& LoRA     
  & \textbf{89.13} & \textbf{72.01} & \textbf{48.59} & \textbf{82.12} 
  & \textbf{93.32} & \textbf{79.83} & \textbf{48.78} & \textbf{87.51} \\

& Qwen2.5 & 3B & VisualPrefix 
  & 65.17 & 36.36 & 34.43 & 54.67 
  & 78.01 & 51.38 & 40.72 & 69.72 \\

\rowcolor{gray!15}
& Qwen2.5 & 3B & Partial Unfreeze \& LoRA     
  & \textbf{71.96} & \textbf{44.61} & \textbf{43.31} & \textbf{59.88} 
  & \textbf{91.25} & \textbf{74.65} & \textbf{47.65} & \textbf{77.85} \\

& StableLM Zephyr & 3B & VisualPrefix 
  & 75.69 & 53.91 & 45.17 & 70.10
  & 61.18 & 29.05 & 30.21 & 53.25 \\

\rowcolor{gray!15}
& StableLM Zephyr & 3B & Partial Unfreeze \& LoRA     
  & \textbf{90.49} & \textbf{74.40} & \textbf{48.13} & \textbf{83.24} 
  & \textbf{75.80} & \textbf{46.73} & \textbf{38.33} & \textbf{57.88} \\
  
& LLaMA 3.2 & 3B & VisualPrefix 
  & 86.78 & 68.66 & 44.62 & 78.70 
  & 83.34 & 65.42 & 45.91 & 72.01 \\

\rowcolor{gray!15}
& LLaMA 3.2 & 3B & Partial Unfreeze \& LoRA     
  & \textbf{93.41} & \textbf{79.61} & \textbf{49.82} & \textbf{88.01}
    & \textbf{91.89} & \textbf{75.88} & \textbf{47.53} & \textbf{83.66} \\
\bottomrule
\end{tabular}
\end{table*}

\section{Results}
\label{sec:results}

We evaluate \textbf{AeroLite} on two remote sensing caption datasets, \textbf{UCM Captions} and \textbf{Sydney Captions}, comparing against both classic CNN+RNN baselines and modern large-scale vision–language models. Table \ref{tab:cnn_vlm_fulltune} provides an overview of these baselines, while Tables \ref{tab:visualprefix_vs_partial_lora} and \ref{tab:tag_vs_notag} report key ablation studies on partial unfreezing vs.\ MLP-only tuning, as well as the role of multi-label tags.

\subsection{Comparison with Baselines}
Despite the sophistication of larger VLMs (7B–13B parameters), our relatively compact \textbf{3B}-scale LLM under \textbf{AeroLite} achieves competitive or superior results on both UCM and Sydney datasets. On UCM, legacy CNN+RNN pipelines (e.g., Show \& Tell, Hard-attention) typically plateau around 50--60\% BLEU-4, while more modern VLMs push towards 60--70\%. In contrast, AeroLite attains 79.61\% BLEU-4 and 88.01\% ROUGE-L, surpassing many heavier models by over 10 points. Likewise, on Sydney Caption Dataset, we record 75.88\% BLEU-4—again exceeding strong 7B--13B contenders. These gains highlight two key factors in our design: \emph{(i)} explicit multi-label tag guidance, which provides specialized domain cues for the language model; and \emph{(ii)} partial unfreezing (plus LoRA), enabling the model to adapt effectively despite its smaller parameter count. By contrast, older CNN+RNN pipelines and even some modern VLMs either lack aerial-specific knowledge or demand immense computational resources. AeroLite not only outperforms these methods in caption fidelity but does so at a fraction of the model size.

\begin{figure*}[!ht]
    \centering
    \begin{subfigure}{0.45\linewidth}
        \centering
        \includegraphics[width=\linewidth]{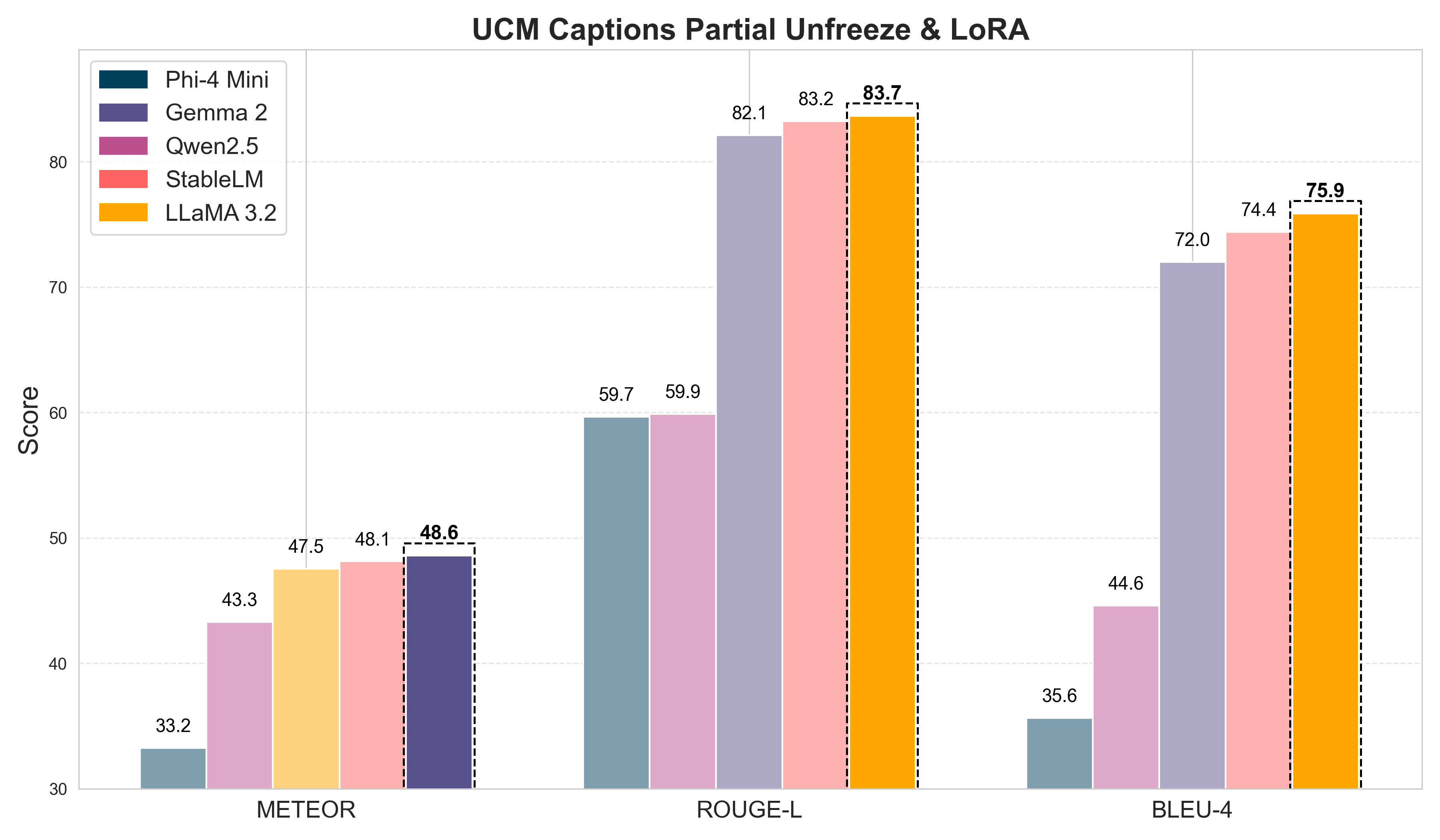}
        \caption{\textbf{UCM Captions Partial Unfreeze \& LoRA}}
        \label{fig:ucm_lora}
    \end{subfigure}\hfill
    \begin{subfigure}{0.45\linewidth}
        \centering
        \includegraphics[width=\linewidth]{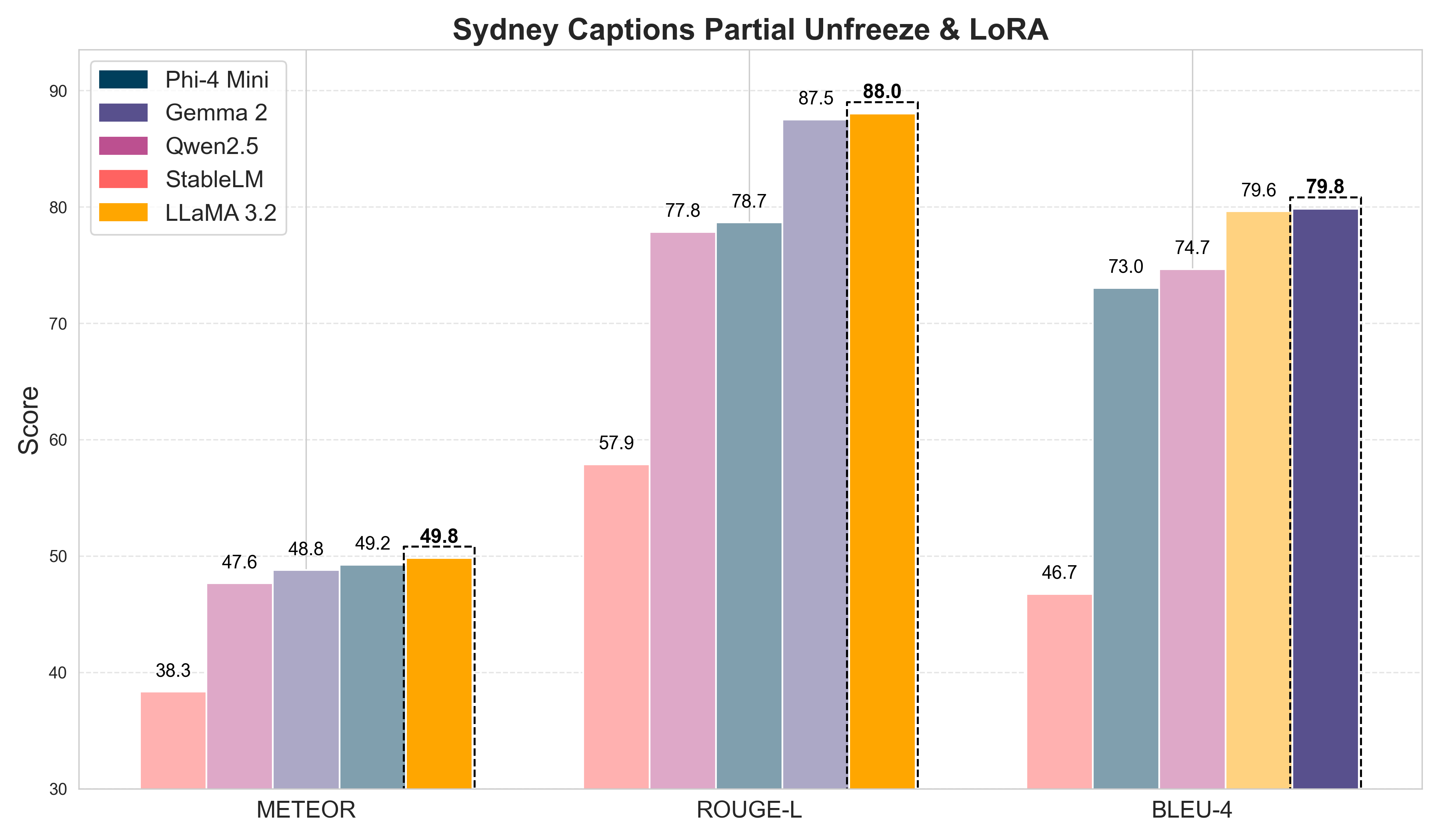}
        \caption{\textbf{Sydney Captions Partial Unfreeze \& LoRA}}
        \label{fig:sydney_lora}
    \end{subfigure}
    \caption{Bar-chart visualization of partial unfreeze \& LoRA performance across different models on the UCM and Sydney caption datasets. 
    In each metric group (BLEU-4, METEOR, ROUGE-L), the bars are sorted from lower to higher, and the best-performing model is highlighted by a dashed rectangle.}
    \label{fig:lora_charts}
\end{figure*}

\subsection{Ablation Study}
\subsubsection{VisualPrefix (MLP Only) vs. Partial Unfreeze \& LoRA.} o better understand AeroLite’s effectiveness, we first investigate how different strategies for visual-language alignment affect captioning performance. In Table~\ref{tab:visualprefix_vs_partial_lora}, we compare two integration approaches across multiple small-scale language models (Phi-4 Mini, Gemma 2, Qwen2.5, StableLM Zephyr, LLaMA 3.2):

\begin{itemize} 
\item \textbf{VisualPrefix (MLP Only):} A lightweight bridging module (MLP) integrates visual embeddings from CLIP directly into the frozen LLM, leaving the entire LLM backbone unchanged. 
\item \textbf{Partial Unfreeze \& LoRA:} We strategically unfreeze and fine-tune the top 30\% of layers using LoRA, allowing the model to directly adapt its internal parameters to aerial-domain semantics. 
\end{itemize}

Our results consistently demonstrate that the Partial Unfreeze \& LoRA approach substantially outperforms the simpler VisualPrefix method, yielding BLEU-4 score increases ranging from approximately +5 to +15 points. Specifically, when applying LLaMA~3.2 to the UCM dataset, the Partial Unfreeze \& LoRA approach dramatically enhances BLEU-4 from 68.66\% (MLP-only) to 79.61\%, reflecting a considerable improvement of nearly 11 percentage points. Similar significant improvements are observed with other models: for instance, StableLM Zephyr increases from 53.91\% to 74.40\%, and Gemma 2 improves from 60.92\% to 72.01\%. This robust pattern underscores that targeted parameter adaptation within the language model is vital for effectively modeling the complex spatial semantics unique to aerial imagery.

\subsubsection{Effectiveness of Multi-Label Tagging.} Next, we examine the role of explicit multi-label \emph{tags} in guiding the caption generation process. In Table~\ref{tab:tag_vs_notag}, we present a direct comparison between AeroLite setups with and without tags provided by the CLIP encoder. The inclusion of these semantic tags results in dramatic performance gains. Specifically, BLEU-4 scores on UCM increase from 61.15\% without tags to 79.61\% with tags, and similarly on Sydney from 43.03\% to 75.88\%.

As shown on ~\ref{fig:aerolite_taggedsamples}, this substantial improvement confirms that tags effectively function as explicit domain-specific guidance for the language model, emphasizing key aerial-scene elements such as \emph{"industrial complex," "airport runway," "agricultural field," and "residential area."} By explicitly incorporating these semantic cues, AeroLite successfully mitigates hallucinations and ensures richer, more accurate, and domain-aware scene descriptions.

\begin{table}[!h]
\centering
\caption{Ablation: \textbf{Partial Unfreeze \& LoRA} w/o Tag vs.\ w/ Tag on UCM and Sydney in LLaMA3.2.}
\label{tab:tag_vs_notag}
\small
\begin{tabular}{lcccccc}
\toprule
& \multicolumn{3}{c}{\textbf{UCM}} & \multicolumn{3}{c}{\textbf{Sydney}} \\
\cmidrule(lr){2-4}\cmidrule(lr){5-7}
\textbf{Metric} & w/o Tag & w/ Tag & \quad & w/o Tag & w/ Tag & \\
\midrule
BLEU-1   & 85.70  & 93.41  && 76.03  & 91.89  & \\
BLEU-2   & 76.44  & 88.20  && 62.74  & 86.56  & \\
BLEU-3   & 68.19  & 83.56  && 52.19  & 80.91  & \\
BLEU-4   & 61.15  & 79.61  && 43.03  & 75.88  & \\
METEOR   & 43.59  & 49.82  && 34.99  & 47.53  & \\
ROUGE-L  & 75.94  & 88.01  && 63.29  & 83.66  & \\
\bottomrule
\end{tabular}
\end{table}

Taken together, these experiments illustrate a clear progression from a simpler visual bridging approach (MLP-only) to targeted language-model adaptation via partial unfreezing and LoRA, further enhanced by explicit multi-label tags. Remarkably, despite relying on a modest 3B-parameter LLM, \textbf{AeroLite} consistently surpasses classical CNN-based methods and matches or outperforms state-of-the-art vision–language frameworks using significantly larger models. These results demonstrate the effectiveness of combining lightweight adaptation techniques and domain-specific semantic guidance, showcasing AeroLite as an efficient, robust, and scalable approach to specialized remote sensing image captioning.

\section{Limitations}
\label{sec:limitations}

Despite the competitive performance of \textbf{AeroLite} using relatively compact language models (1–3B parameters), it still has several notable limitations. First, the reliance on pseudo-captions generated from polygon annotations may introduce positional inaccuracies or omit finer details, especially when GPT-generated prompts lack adequate context. Maintaining consistent data quality across diverse aerial datasets remains a challenge. Second, while multi-label tagging provides valuable domain-specific cues, any errors or inaccuracies in these tags can directly propagate into the language model, potentially causing extraneous or misleading descriptions. Additionally, due to its small scale, the language model inherently faces challenges such as inevitable hallucinations. It might overemphasize minor details or misinterpret certain objects—and significantly weakened continuous conversational context abilities.

Third, although the partial unfreeze and LoRA strategies substantially reduce computational costs compared to full fine-tuning, they still necessitate specialized GPU resources. Hence, deploying AeroLite in extremely resource-constrained edge environments, such as drones with minimal computational capabilities, may require additional optimization strategies like pruning, quantization, or further model distillation.

\section{Conclusion}
\label{sec:conclusion}

We have presented \textbf{AeroLite}, a tag-guided remote sensing captioning framework that effectively integrates CLIP-based multi-label classification, a compact bridging MLP module, and a partial unfreeze and LoRA tuning strategy to adapt modestly sized language models. Through extensive evaluations on benchmark datasets such as UCM and Sydney, AeroLite not only surpasses classical CNN-based methods but also demonstrates competitive performance compared to larger-scale vision-language models. Central to AeroLite's success is the strategic use of explicit semantic tagging and targeted layer adaptation, enabling even a relatively small-scale language model (approximately 3B parameters) to accurately interpret complex aerial scenes.

Despite its notable strengths, AeroLite has inherent limitations due to its model size, including occasional hallucinations and weaker sustained contextual reasoning capabilities. Moving forward, several promising development avenues emerge. Incorporating higher-resolution imagery and integrating time-series data could significantly enhance the model's ability to capture dynamic seasonal or temporal variations. Additionally, combining object detection and captioning into a unified framework could facilitate simultaneous object discovery and rich textual scene descriptions. Exploring cross-task learning, such as integrating scene segmentation or change detection alongside captioning, would further leverage synergistic improvements in remote sensing analyses.

By continually refining multi-label tagging strategies, partial unfreezing methodologies, and domain-adaptive prompts, we aim to mitigate hallucinations, enhance contextual understanding, and further elevate AeroLite’s capability to generate accurate, contextually detailed aerial scene interpretations. Ultimately, AeroLite serves as a foundational step toward broader multi-task aerial intelligence systems, bridging captioning, detection, segmentation, and beyond, particularly in resource-constrained remote sensing applications.

\bibliographystyle{ACM-Reference-Format}
\bibliography{reference}
\end{document}